\def\ARXIV{1} 
\def\PAGENUMS{0} 
\def\AAM{0} 
\ifnum\ARXIV=1 
    \def\PAGENUMS{1} 
    \def\AAM{1} 
\else
    \def\PAGENUMS{0} 
    \def\AAM{0} 
\fi

\documentclass[conference]{IEEEtran}
\IEEEoverridecommandlockouts
\usepackage{booktabs}
\usepackage{multirow}
\usepackage{threeparttable}
\usepackage{caption}
\usepackage{siunitx} 
\usepackage[normalem]{ulem}
\usepackage{cite}
\usepackage{amsmath,amssymb,amsfonts}
\usepackage{mathtools}
\usepackage{algorithmic}
\usepackage{graphicx}
\usepackage{textcomp}
\usepackage{xcolor}

\ifnum\PAGENUMS=1
    \usepackage{fancyhdr} 
    \usepackage{lastpage} 
\fi

\newcommand{\boldredres}[1]{\textcolor{red}{\textbf{#1}}}
\newcommand{\secondblueres}[1]{\textcolor{blue}{\underline{#1}}}
\newcommand{\ms}[2]{#1\,$\pm$\,#2}

\usepackage[hidelinks]{hyperref}
\def\BibTeX{{\rm B\kern-.05em{\sc i\kern-.025em b}\kern-.08em
    T\kern-.1667em\lower.7ex\hbox{E}\kern-.125emX}}
\begin{document}

\title{VMDNet: Temporal Leakage-Free Variational Mode Decomposition for Electricity Demand Forecasting%
\thanks{W. Feng's PhD research is supported by the UKRI EPSRC Doctoral Training Partnership (EP/W524414/1). J. Cartlidge is supported by UKRI EPSRC Grant EP/Y028392/1: AI for Collective Intelligence (AI4CI).}
}


\author{
\IEEEauthorblockN{Weibin Feng$^{1}$\qquad
Ran Tao$^{2}$\qquad
John Cartlidge$^{1}$\qquad
Jin Zheng$^{1}$}

\IEEEauthorblockA{$^{1}$ School of Engineering Mathematics and Technology, University of Bristol, UK \\
$^{2}$ University of Bristol Business School, University of Bristol, UK}
}

\maketitle

\ifnum\PAGENUMS=1
    \thispagestyle{fancy}
    \pagestyle{fancy}
    \fancyfoot[C]{\fontsize{8}{10} \selectfont Page \thepage ~of {\hypersetup{hidelinks}\pageref{LastPage}}}
    \fancyhead[L,C,R]{} 
    \setlength{\headheight}{10.0pt}
    \ifnum\AAM=1 
        \fancyhead[C]{\fontsize{8}{10} \selectfont Accepted author manuscript: 
        Feng et al. (2026), 34th European Signal Processing Conference (EUSIPCO), Bruges, Belgium.}
    \else
        \renewcommand{\headrulewidth}{0pt} 
    \fi
\fi

\begin{abstract}
Accurate electricity demand forecasting is challenging due to the strong multi-periodicity of real-world demand series, which makes effective modeling of recurrent temporal patterns crucial. Decomposition techniques make such structure explicit and thereby improve predictive performance. Variational Mode Decomposition (VMD) is a powerful signal-processing method for periodicity-aware decomposition and has seen growing adoption in recent years. However, existing studies often suffer from information leakage and rely on inappropriate hyperparameter tuning. To address these issues, we propose VMDNet, a causality-preserving framework that (i) applies sample-wise VMD to avoid temporal leakage; (ii) represents each decomposed mode with frequency-aware embeddings and decodes it using parallel temporal convolutional networks (TCNs), ensuring mode independence and efficient learning; and (iii) introduces a Stackelberg game inspired bilevel scheme to guide the selection of VMD's two key hyperparameters. Experiments on three widely used electricity demand datasets show that VMDNet consistently outperforms state-of-the-art baselines. 
\end{abstract}


\section{Introduction}
\noindent
Accurate electricity demand forecasting is a fundamental task for the reliable and economical operation of modern power systems \cite{DONG2025110980}. It directly supports generation scheduling, grid dispatch, and market clearing, and becomes increasingly challenging with the growing penetration of renewable energy sources\cite{WU2021683}. Given the inherent dynamics and temporal dependencies of electric load data \cite{HAQUE2022108877}, accurate modeling of periodic patterns and multi-scale structures is critical for robust forecasting \cite{Zhou2021Informer:Forecasting}. This motivates the incorporation of decomposition techniques, which aim to separate raw sequences into interpretable components that facilitate predictions.


VMD \cite{Dragomiretskiy2014VariationalDecomposition} is an advanced signal processing technique that decomposes a time series into a finite number of band-limited modes, also called intrinsic mode functions (IMFs), each centered on a distinct frequency. Unlike traditional tools such as Fourier or wavelet transforms, VMD enforces spectral sparsity and mode separation through variational optimization, producing non-overlapping modes with compact bandwidths. This leads to a more interpretable and disentangled representation of the original signal. Empirical evidence has demonstrated that it is effective in energy forecasting \cite{MA2025124246, han_research_2025}.

However, two issues remain and potentially prevent its practical application to forecasting: (i) the prevalent practice of decomposing the entire series before training grants implicit access to future information and violates causality~\cite{QIAN2019939}; and (ii) hyperparameter sensitivity, such that performance depends critically on the number of modes $K$ and the bandwidth penalty $\alpha$ \cite{KRISHNARAYI2022122585}. 

 Regarding issue (i), some studies decompose the data after the train/test split to avoid cross-set leakage \cite{CHEN2024107353,HUANG2021116485}. Even so, the decomposition is still computed on entire input subsequences, which leaks future information. Moreover, short-window decomposition often yields weakly separated modes whose identities drift across samples, and the forecasting architectures used in these works are not designed to robustly represent such unstable modes.
 
 For issue (ii), existing works mostly rely on meta-heuristic searches (e.g., GA, PSO) to optimise simple proxy measures or statistics, such as reconstruction error or sample entropy to raise decomposition quality \cite{choudhary_genetic_2025,AN20228574}. Nevertheless, gains under these proxies do not necessarily translate into better forecasting, because such proxies ignore the representational requirements of the downstream predictor. In addition, most previous studies typically treat $K$ and $\alpha$ independently, despite their asymmetric importance.

To address these limitations in a unified and task-aware manner, we propose VMDNet, a leakage-free, multi-branch forecasting framework built upon variational mode decomposition. Our main contributions are summarised as follows.
\begin{itemize}
  \item We propose a sample-wise VMD scheme to eliminate temporal information leakage in forecasting settings.
  \item Our framework employs a multi-branch decoding architecture with frequency-aware embeddings to model decomposed modes independently and robustly.
  \item A bilevel search motivated by the Stackelberg game is explored to provide task-aware guidance for selecting key VMD hyperparameters.
\end{itemize}

The implementation is released with a GPU-parallelised batch VMD module.\footnote{\url{https://github.com/weibin-feng/VMDNet}}

\section{Methodology}

\label{sec:method}

\begin{figure*}[tbp]
    \centering
    \includegraphics[width=0.95\linewidth,trim=5 5 5 5, clip]{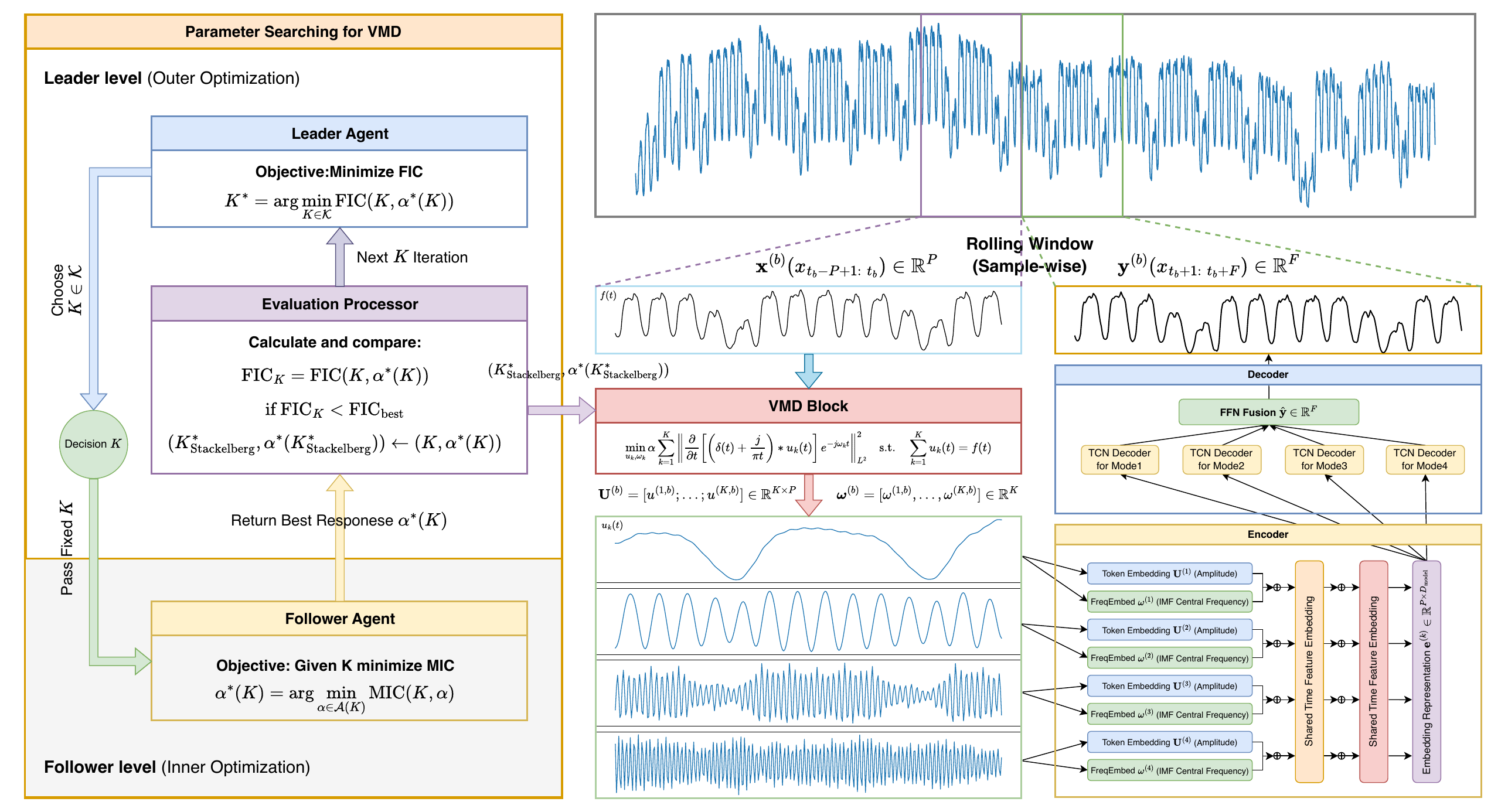}
    \caption{Schematic of VMDNet. An example configuration on the ENTSO-E dataset with $K=4$ is shown.}

    \label{fig:mechanism}
\end{figure*}

\subsection{Variational Mode Decomposition}
\label{sec:vmd_methodology}

\noindent
VMD \cite{Dragomiretskiy2014VariationalDecomposition} 
decomposes a signal $f(t)$ into $K$ band-limited IMFs $u_k(t)$, 
each centred at a frequency $\omega_k$, by solving a constrained variational optimisation problem:
\begin{equation}
\min_{\{u_k\}, \{\omega_k\}} \left\{ \alpha\,\sum_{k=1}^{K} 
\big\| \partial_t \big[\big( \delta(t) + \tfrac{j}{\pi t} \big) * u_k(t)\big] e^{-j\omega_k t} \big\|_{L^2}^2 \right\}
\label{eq:original_vmd_time}
\end{equation}
subject to the reconstruction constraint
\begin{equation}
\sum_{k=1}^{K} u_k(t) = f(t).
\label{eq:constraint_time}
\end{equation}
Here, $\alpha>0$ penalises the mode bandwidth, $*$ denotes the convolution, and $j^2=-1$. 
The Hilbert transform $(\delta(t)+\tfrac{j}{\pi t})*u_k(t)$ produces the analytic signal of $u_k(t)$; 
multiplying by $e^{-j\omega_k t}$ shifts it to baseband; 
$\partial_t$ then measures the variation of this demodulated signal, 
and the $L^2$ norm quantifies the bandwidth around $\omega_k$. 
The optimisation in Eqs.~\eqref{eq:original_vmd_time}--\eqref{eq:constraint_time} is solved via ADMM.

\subsection{Sample-wise Decomposition} \label{sec:samplewise} 

\noindent
Let the time series be $\mathcal{D}=\{x_t\}_{t=1}^{T}$. Given lookback length $P$ and forecast horizon $F$, the forecasting problem is to learn a mapping \begin{equation} \mathcal{F}_\theta:\;\mathbb{R}^{P}\!\to\!\mathbb{R}^{F},\quad \hat{\mathbf{y}}^{(b)} = \mathcal{F}_\theta(\mathbf{x}^{(b)}), \label{eq:forecast_map} \end{equation} where $\mathbf{x}^{(b)}\in\mathbb{R}^{P}$ is the $b$-th input window and $\hat{\mathbf{y}}^{(b)}\in\mathbb{R}^{F}$ is the corresponding prediction. 

For a stride $s\!\ge\!1$, define the endpoint of the $b$-th window as $t_b = P + (b-1)s$. The input--target pair is then 
\begin{equation}
\begin{aligned}
\mathbf{x}^{(b)} &\coloneqq (x_{t_b-P+1:\,t_b}) \in \mathbb{R}^{P}, \\
\mathbf{y}^{(b)} &\coloneqq (x_{t_b+1:\,t_b+F}) \in \mathbb{R}^{F}.
\end{aligned}
\label{eq:rolling_window}
\end{equation}

Stacking across $b=1,\dots,B$ windows yields matrices $\mathbf{X}\in\mathbb{R}^{B\times P}$ and $\mathbf{Y}\in\mathbb{R}^{B\times F}$. 

Each input window $\mathbf{x}^{(b)}$ is decomposed by a $K$-mode VMD operator with
bandwidth penalty $\alpha$:
\begin{align}
\mathbf{U}^{(b)} &= [u^{(1,b)};\dots;u^{(K,b)}] \in \mathbb{R}^{K\times P}, \\
\boldsymbol{\omega}^{(b)} &= [\omega^{(1,b)},\dots,\omega^{(K,b)}] \in \mathbb{R}^{K}, 
\end{align}
where $u^{(k,b)}$ denotes the $k$-th mode and $\omega^{(k,b)}$ its estimated center frequency. Each window satisfies the reconstruction constraint $\mathbf{x}^{(b)} = \sum_{k=1}^{K} u^{(k,b)}$.

Stacking across $b$ gives
\begin{equation}
\mathbf{U}\in\mathbb{R}^{B\times K\times P},\quad
\boldsymbol{\Omega}\in\mathbb{R}^{B\times K}.
\label{eq:samplewise_stack}
\end{equation}

The decomposition on $\mathbf{x}^{(b)}$ depends only on 
$(x_{t_b-P+1:\;t_b})$, i.e., observations available at time $t_b$. 
Thus, samplewise decomposition preserves causality and avoids information leakage by design.

\subsection{Encoding Architecture}
\label{sec:embedding}

\noindent
Instead of applying heavy sequence encoders, we exploit the clean spectral 
structure of IMFs and adopt a simple embedding strategy as the encoder. For each sample $b=1,\dots,B$, mode $k=1,\dots,K$, and timestep $t=1,\dots,P$, 
the embedded representation $\mathbf{e}_t^{(b,k)} \in \mathbb{R}^{d_{\text{model}}}$ is
\begin{align}
\mathbf{e}_t^{(b,k)} &= 
    \underbrace{\text{TokenEmbed}_k\!\left(u_t^{(k,b)}\right)}_{\text{amplitude}}
    + \underbrace{\text{TimeEmbed}(t) + \text{PosEmbed}(t)}_{\text{shared}} \nonumber \\
    &\quad + \underbrace{\text{FreqEmbed}_k\!\left(\omega^{(k,b)}\right)}_{\text{frequency}}.
\label{eq:embedding}
\end{align}

\noindent The embedding components are defined as follows:
\begin{itemize}
    \item $\text{TokenEmbed}_k: \mathbb{R}\!\to\!\mathbb{R}^{d_{\text{model}}}$  
    is implemented as a single-layer 1D convolution 
(kernel size 3 with circular padding) projecting the input sequence into 
$\mathbb{R}^{d_{\text{model}}}$ to capture local amplitude patterns.

    \item $\text{FreqEmbed}_k: \mathbb{R}\!\to\!\mathbb{R}^{d_{\text{model}}}$  
    projects the center frequency from Eq.~\eqref{eq:samplewise_stack} into the embedding space
    via a learnable linear layer mapping the scalar center frequency to 
$\mathbb{R}^{d_{\text{model}}}$.

\item $\text{TimeEmbed}, \text{PosEmbed}: \mathbb{N}\!\to\!\mathbb{R}^{d_{\text{model}}}$  
are, respectively, a linear projection of calendar-time features and fixed sinusoidal positional encodings, shared across all modes.
\end{itemize}

Stacking across all samples, modes, and timesteps yields the representation tensor
\begin{equation}
\mathbf{E} \in \mathbb{R}^{B \times K \times P \times d_{\text{model}}},
\label{eq:encoder_output}
\end{equation}
This embedding design introduces an explicit frequency domain inductive bias into the model, allowing each mode to carry knowledge of its spectral identity throughout the learning pipeline. Unlike standard token embeddings which focus solely on amplitudes, the proposed structure respects both temporal and spectral semantics of the decomposed signals.

To capitalise on the decomposition, we treat each mode as a distinct frequency scale and model it separately. Concretely, we use mode specific parameterisations for \(\text{TokenEmbed}_k\) and \(\text{FreqEmbed}_k\), while sharing \(\text{TimeEmbed}\) and \(\text{PosEmbed}\) to preserve global temporal alignment. This specialisation preserves sub-frequency characteristics and mitigates cross mode interference (mode mixing), whereas globally shared weights would force heterogeneous sub-bands into a single representation and dilute mode specific information.

\subsection{TCN-Based Parallel Decoding}
\label{sec:tcn-decoder}

\noindent
We used a parallel decoding mechanism that assigns one TCN branch per mode. 
For mode $k$, the slice $\mathbf{E}^{(k)}\in\mathbb{R}^{B\times P\times d_{\text{model}}}$ 
is decoded as
\begin{equation}
\hat{\mathbf{y}}^{(k)} = \mathrm{TCN}_k(\mathbf{E}^{(k)}) \in \mathbb{R}^{B\times F\times 1}.
\label{eq:tcn_mode}
\end{equation}
Each $\mathrm{TCN}_k(\cdot)$ consists of $L$ residual blocks with dilated 
convolutions ($d=2^\ell,\;\ell=0\dots L-1$), causal padding, GELU activations, 
and dropout. This preserves causality and captures long-range dependencies. Outputs from all branches are stacked as
\begin{equation}
\hat{\mathbf{Y}}=\mathrm{stack}\big(\hat{\mathbf{y}}^{(1)},\dots,\hat{\mathbf{y}}^{(K)}\big)
\in \mathbb{R}^{B\times K\times F\times 1}.
\label{eq:decoder_output}
\end{equation}
Finally, a lightweight feedforward network (FFN) aggregates mode-wise forecasts 
into the final prediction $\hat{\mathbf{y}}\in\mathbb{R}^{B\times F\times 1}$.

Symmetrically, the parallel decoding design follows the same principle of
modeling each frequency scale separately. For strongly periodic, multi-scale
series, sub-band signals are typically more predictable and locally
stationary, so each branch can employ a lightweight causal decoder (dilated
TCN) to achieve high accuracy with small capacity. Running \(K\) such decoders
in parallel preserves the spectral separability of VMD modes, mitigates
cross-mode interference, and yields efficient causal modeling. Finally, we fuse
the branch-wise outputs with a two-layer MLP to produce the final forecast,
thereby completing the decoding stage.

\subsection{Parameter Searching for VMD}
\label{sec:vmd_parameter_selection}

\noindent
VMD requires tuning two key parameters: the number of modes $K$ and the bandwidth penalty $\alpha$.
These parameters play asymmetric roles: $K$ sets the decomposition resolution (dominant),
whereas $\alpha$ refines spectral compactness within that resolution (secondary).
To address this asymmetry, we adopt a Stackelberg game process. A Stackelberg game is a hierarchical framework with two agents in unequal roles: a leader moves first, anticipating the follower's best response; the follower observes the leader's decision and solves a conditional problem to respond optimally.
The resulting outcome is a Stackelberg equilibrium, where neither side benefits from deviating given this sequence of moves \cite{VonStackelberg2011MarketEquilibrium,Colson2007AnOptimization}. 

 


Instantiated for VMD, we define two objectives: (i) the Forecastability Information Criterion (FIC), which quantifies forecastability by fitting an $AR(r)$ model to each mode and summing the residual variances; and (ii) the Mutual Information Criterion (MIC), which quantifies cross-mode overlap via average pairwise mutual information. FIC additionally includes an AIC/BIC-inspired complexity penalty to discourage over-granular decompositions. Formally,

\begin{equation}
\label{eq:fic}
\begin{aligned}
\mathrm{FIC}(K,\alpha) &= (T{-}r)\,\log\!\Big(\sum_{k=1}^{K}\hat{\sigma}_k^2(K,\alpha)\Big) \\
&\quad + \big(K(r{+}1){+}1\big)\,\log(T{-}r).
\end{aligned}
\end{equation}

\begin{equation}
\label{eq:mic}
\mathrm{MIC}(K,\alpha) = \frac{2}{K(K{-}1)} \sum_{i=1}^{K-1}\sum_{j=i+1}^{K} \hat{I}(u_i; u_j; K, \alpha).
\end{equation}

Let $\mathcal{K}=\{K_{\min},\dots,K_{\max}\}$ and $\mathcal{A}(K)=[\alpha_{\min},\alpha_{\max}]$, this yields a
Stackelberg bilevel search in Eqs.~\eqref{eq:stkbg_alpha}--\eqref{eq:stkbg_K},
\begin{align}
\alpha^*(K) &= \arg\min_{\alpha\in\mathcal{A}(K)} \mathrm{MIC}(K,\alpha),
\label{eq:stkbg_alpha}\\
K^* &= \arg\min_{K\in\mathcal{K}} \mathrm{FIC}\big(K,\alpha^*(K)\big).
\label{eq:stkbg_K}
\end{align}

Since the optimisation is obtained via a heuristic without formal convergence guarantees, we perform multiple independent restarts with different random seeds, evaluate the resulting $(K,\alpha)$ pairs on the validation set, and select the best-performing pair as $(K^\star,\alpha^\star)$.

\begin{table*}[htbp]
  \caption{Forecasting results at horizons $\{96, 192, 336\}$ with a lookback length of 336. Results are averaged over 5 runs with seeds $\{2021, 2022, 2023, 2024, 2025\}$. Each cell reports test MSE and MAE (mean $\pm$ std.); lower is better. Best and second best results are highlighted in \boldredres{red bold} and \secondblueres{blue underline}, respectively.\label{forecastingresults}}
  \centering
  \renewcommand{\arraystretch}{1.55}
  \resizebox{1.0\textwidth}{!}{%
    \begin{threeparttable}
      \fontsize{10pt}{8pt}\selectfont
      \renewcommand{\multirowsetup}{\centering}
      \setlength{\tabcolsep}{2pt}
      \begin{tabular}{c|c|ccc|ccc|ccc}
        \toprule
        \multirow{2}{*}{\textbf{Models}} & \multirow{2}{*}{\textbf{Metric}} &
        \multicolumn{3}{c}{\textbf{ENTSO-E}} &
        \multicolumn{3}{c}{\textbf{ISO-NE}} &
        \multicolumn{3}{c}{\textbf{AEMO}} \\
        \cmidrule(lr){3-5} \cmidrule(lr){6-8} \cmidrule(lr){9-11}
         &  & 96 & 192 & 336 & 96 & 192 & 336 & 96 & 192 & 336 \\
        \midrule

        \multirow{2}{*}{\textbf{VMDNet}}
          & MSE & \boldredres{\ms{0.156}{0.011}} & \boldredres{\ms{0.217}{0.015}} & \boldredres{\ms{0.231}{0.007}}
                & \secondblueres{\ms{0.196}{0.009}} & \boldredres{\ms{0.222}{0.007}} & \boldredres{\ms{0.216}{0.016}}
                & \boldredres{\ms{0.369}{0.005}} & \boldredres{\ms{0.414}{0.020}} & \boldredres{\ms{0.408}{0.005}} \\
          & MAE & \boldredres{\ms{0.261}{0.007}} & \boldredres{\ms{0.304}{0.012}} & \boldredres{\ms{0.322}{0.011}}
                & \secondblueres{\ms{0.344}{0.010}} & \boldredres{\ms{0.369}{0.006}} & \boldredres{\ms{0.363}{0.013}}
                & \secondblueres{\ms{0.426}{0.005}} & \boldredres{\ms{0.455}{0.012}} & \boldredres{\ms{0.453}{0.005}} \\
        \midrule

        \multirow{2}{*}{TimeMixer \cite{Wang2024TIMEMIXER:FORECASTING}}
          & MSE & \ms{0.173}{0.012} & \ms{0.230}{0.008} & \ms{0.287}{0.016}
                & \ms{0.200}{0.005} & \ms{0.238}{0.022} & \ms{0.225}{0.011}
                & \ms{0.411}{0.030} & \ms{0.439}{0.014} & \secondblueres{\ms{0.444}{0.023}} \\
          & MAE & \ms{0.275}{0.009} & \ms{0.328}{0.008} & \ms{0.372}{0.009}
                & \ms{0.348}{0.005} & \ms{0.380}{0.018} & \secondblueres{\ms{0.368}{0.010}}
                & \ms{0.439}{0.018} & \ms{0.467}{0.008} & \ms{0.473}{0.017} \\
        \midrule

        \multirow{2}{*}{iTransformer \cite{Liu2024ITRANSFORMER:FORECASTING}}
          & MSE & \ms{0.182}{0.005} & \ms{0.229}{0.010} & \ms{0.402}{0.050}
                & \ms{0.201}{0.005} & \ms{0.233}{0.003} & \ms{0.240}{0.005}
                & \ms{0.381}{0.010} & \ms{0.460}{0.016} & \ms{0.542}{0.008} \\
          & MAE & \ms{0.290}{0.003} & \ms{0.330}{0.008} & \ms{0.447}{0.035}
                & \ms{0.347}{0.005} & \ms{0.379}{0.003} & \ms{0.386}{0.004}
                & \ms{0.433}{0.007} & \ms{0.489}{0.011} & \ms{0.542}{0.006} \\
        \midrule

        \multirow{2}{*}{Crossformer \cite{Zhang2023Crossformer:Forecasting}}
          & MSE & \ms{0.277}{0.074} & \ms{0.430}{0.131} & \ms{0.555}{0.043}
                & \ms{0.206}{0.007} & \ms{0.253}{0.049} & \ms{0.258}{0.043}
                & \ms{0.382}{0.015} & \ms{0.442}{0.012} & \ms{0.482}{0.006} \\
          & MAE & \ms{0.363}{0.078} & \ms{0.488}{0.119} & \ms{0.580}{0.024}
                & \ms{0.353}{0.007} & \ms{0.395}{0.041} & \ms{0.400}{0.034}
                & \ms{0.431}{0.012} & \ms{0.473}{0.010} & \ms{0.498}{0.004} \\
        \midrule

        \multirow{2}{*}{DLinear \cite{Zeng2023AreForecasting}}
          & MSE & \ms{0.220}{0.002} & \ms{0.267}{0.001} & \ms{0.334}{0.005}
                & \ms{0.198}{0.001} & \ms{0.239}{0.000} & \ms{0.223}{0.000}
                & \ms{0.375}{0.000} & \secondblueres{\ms{0.422}{0.000}} & \ms{0.463}{0.000} \\
          & MAE & \ms{0.300}{0.003} & \ms{0.328}{0.001} & \ms{0.378}{0.006}
                & \secondblueres{\ms{0.344}{0.000}} & \ms{0.378}{0.001} & \ms{0.371}{0.002}
                & \boldredres{\ms{0.417}{0.000}} & \secondblueres{\ms{0.460}{0.000}} & \secondblueres{\ms{0.470}{0.000}} \\
        \midrule

        \multirow{2}{*}{PatchTST \cite{Nie2023ATRANSFORMERS}}
          & MSE & \ms{0.167}{0.003} & \ms{0.226}{0.006} & \ms{0.286}{0.004}
                & \ms{0.211}{0.021} & \ms{0.246}{0.010} & \ms{0.278}{0.005}
                & \ms{0.378}{0.004} & \ms{0.459}{0.006} & \ms{0.503}{0.014} \\
          & MAE & \secondblueres{\ms{0.267}{0.002}} & \ms{0.314}{0.006} & \ms{0.361}{0.003}
                & \ms{0.355}{0.018} & \ms{0.385}{0.009} & \ms{0.414}{0.004}
                & \ms{0.428}{0.002} & \ms{0.476}{0.003} & \ms{0.502}{0.008} \\
        \midrule

        \multirow{2}{*}{TimesNet \cite{Wu2023TIMESNET:ANALYSIS}}
          & MSE & \secondblueres{\ms{0.163}{0.012}} & \secondblueres{\ms{0.222}{0.035}} & \secondblueres{\ms{0.274}{0.023}}
                & \boldredres{\ms{0.187}{0.004}} & \secondblueres{\ms{0.232}{0.008}} & \secondblueres{\ms{0.220}{0.007}}
                & \secondblueres{\ms{0.372}{0.003}} & \ms{0.431}{0.027} & \ms{0.509}{0.014} \\
          & MAE & \boldredres{\ms{0.261}{0.011}} & \secondblueres{\ms{0.309}{0.025}} & \secondblueres{\ms{0.353}{0.012}}
                & \boldredres{\ms{0.325}{0.003}} & \secondblueres{\ms{0.374}{0.008}} & \secondblueres{\ms{0.368}{0.006}}
                & \ms{0.433}{0.005} & \ms{0.464}{0.017} & \ms{0.507}{0.013} \\

        \bottomrule
      \end{tabular}
    \end{threeparttable}%
  }
\end{table*}

\section{Experiments}
\subsection{Dataset}
\label{ssec:data}

\noindent
We evaluate the proposed model on three widely used electricity demand datasets \cite{DONG2025110980} from (i) Poland power system demand (ENTSO-E)\footnote{\url{https://www.entsoe.eu/data/power-stats/}}, (ii) ISO New England (ISO-NE)\footnote{\url{https://www.iso-ne.com/isoexpress/web/reports/load-and-demand/-/tree/dmnd-rt-hourly-sys}}, and (iii) Australian Energy Market Operator for New South Wales (AEMO)\footnote{\url{https://www.aemo.com.au/energy-systems/electricity/}}. All datasets contain hourly records of average electricity demand and span the period from January 1, 2021 to June 13, 2024 (30{,}216 samples).

\subsection{Baselines}
\noindent
We perform a comprehensive comparison with six recent SOTA time series forecasting models: TimeMixer \cite{Wang2024TIMEMIXER:FORECASTING}, iTransformer \cite{Liu2024ITRANSFORMER:FORECASTING}, Crossformer \cite{Zhang2023Crossformer:Forecasting}, DLinear \cite{Zeng2023AreForecasting}, PatchTST \cite{Nie2023ATRANSFORMERS}, TimesNet \cite{Wu2023TIMESNET:ANALYSIS}. Forecast accuracy is evaluated using mean squared error (MSE) and mean absolute error (MAE).


\subsection{Experiment Setup}
\label{ssec:expset}
\noindent
All models are trained with a fixed lookback length $P=336$ and evaluated at horizons $F\in\{96,192,336\}$. The initial learning rate of $10^{-3}$ with Adam optimizer is adopted, train 10 epochs with early stopping. The batch size is set to 64 and the embedding dimension is 64. For parameter selection, we search $\mathcal{K}=\{2,3,\dots,15\}$ and $\mathcal{A}(K)=[500,\;10000]$, and use AR order \(p=2\) when computing FIC, and perform 20 independent runs. All hyperparameters are tuned on the validation set; minor adjustments across horizons are applied. Training and evaluation are conducted on \(4\times\) NVIDIA GH200 (120 GB) Grace Hopper Superchips (CUDA 12.7) \cite{10.1145/3725789.3725794}.

\subsection{Results and Discussions}
\subsubsection{Results Analysis}
\label{sec:results}
\noindent
As summarised in Table~\ref{forecastingresults}, across 9 experimental settings spanning different datasets and forecasting horizons, VMDNet achieves the best results in 8 settings and the second-best result in the remaining one, demonstrating consistent superiority over strong baselines. Moreover, VMDNet maintains robust performance across datasets from different countries, indicating strong generalisation under diverse demand patterns.

TimesNet \cite{Wu2023TIMESNET:ANALYSIS} achieves the second-best overall performance across the evaluated settings, highlighting the importance of frequency-oriented inductive biases through multi-period unfolding and 2D tensorization that capture periodic structures at a macro level. Performance differences across datasets further reveal how dataset characteristics interact with such inductive biases. ENTSO-E and ISO-NE aggregate demand at the country/region level, yielding more stable periodic patterns, which particularly benefit periodicity-aware models such as TimesNet. In contrast, AEMO only covers New South Wales, where local demand is less averaged and exhibits higher variability, making linear models like DLinear more competitive. Nevertheless, VMDNet remains consistently strong across all datasets. This is because VMD decomposes the signal into band-limited modes via a global optimization objective, with each mode characteised by a well-separated center frequency. Such a frequency domain representation tends to be less sensitive to local noise and short-term fluctuations, as high-frequency noise is largely isolated into high-frequency components or attenuated, yielding more stable frequency embeddings for robust pattern learning. Compared with TimesNet’s high-level periodic modeling, VMDNet explicitly aligns representations in the frequency domain, leading to superior overall performance across diverse demand patterns.

\subsubsection{Ablation Study}
\noindent
We conduct ablation studies to quantify the contribution of each component: (i) removing VMD by forecasting directly from raw inputs with a single TCN decoder; (ii) removing frequency encoding; (iii) removing parallel decoding by mean-aggregating all mode embeddings before a unified TCN decoder; (iv) removing both frequency encoding and parallel decoding; and (v) replacing the proposed bilevel parameter selection scheme with a PSO baseline that minimizes envelope entropy. All variants are trained and evaluated on ENTSO-E with horizon $F{=}336$.

\begin{table}[t]
  \caption{Ablation results on the ENTSO-E dataset.}
  \label{tab:ablation}
  \centering
  \renewcommand{\arraystretch}{0.90}
  \fontsize{8pt}{9pt}\selectfont
  \setlength{\tabcolsep}{12pt}
  \begin{threeparttable}
    \begin{tabular}{c|cc}
      \toprule
\multirow{2}{*}{\textbf{Models / Variants}} & \multicolumn{2}{c}{\textbf{ENTSO-E (336)}} \\
\noalign{\vskip -1.6pt} 
\cmidrule(lr){2-3}
\noalign{\vskip -1.6pt} 
& MSE & MAE \\
\noalign{\vskip -1.6pt} 
\midrule

      \textbf{VMDNet (full)}              & \boldredres{\ms{0.231}{0.007}} & \boldredres{\ms{0.322}{0.011}} \\
      w/o VMD                      & \ms{0.249}{0.004}              & \ms{0.339}{0.006} \\
      w/o frequency encoding            & \ms{0.244}{0.010} & \ms{0.334}{0.002} \\
      w/o parallel decoding             & \ms{0.274}{0.027}              & \ms{0.353}{0.023} \\
      w/o freq. enc. \& par. dec.
           & \ms{0.287}{0.017} & \ms{0.357}{0.007} \\
      w/o bilevel searching        & \secondblueres{\ms{0.240}{0.009}}              & \secondblueres{\ms{0.329}{0.010}} \\
      \bottomrule
    \end{tabular}
  \end{threeparttable}
\end{table}

As shown in Table~\ref{tab:ablation}, compared with the full model, removing VMD results in an MSE increase of 0.018 (7.8\%), highlighting the importance of frequency-aware decomposition. Frequency encoding also contributes consistently, as its removal increases the MSE by 0.013 (5.6\%). Among all components, parallel decoding plays the most critical role, causing the largest performance drop (0.043, 18.6\%) when removed. When both frequency encoding and parallel decoding are discarded, the degradation is further amplified to 0.056 (24.2\%), demonstrating their complementary effects. In contrast, bilevel searching provides a moderate gain (0.009, 3.9\%), serving as a performance booster rather than a dominant factor.

\subsubsection{Computational Cost}
\noindent
The additional cost mainly arises from bilevel parameter search and VMD preprocessing. The bilevel search is performed offline only once before training, taking 556.2 s versus 385.9 s for the PSO baseline while achieving a 3.9\% performance gain. During inference, VMD dominates the overhead (1.7180 ms per sample on average). Nevertheless, due to the lightweight architecture, our method remains efficient, ranking 4th in inference speed among all baselines (1.8519 ms per sample) while achieving consistent gains of up to 7.8\%.

\section{Conclusion}
\label{sec:con}
\noindent
We introduced VMDNet, an information leakage-free VMD-based model for electricity demand forecasting that integrates sample-wise variational mode decomposition with frequency-aware representations and parallel temporal convolutional decoding. By modeling decomposed modes in the frequency domain and guiding hyperparameter selection via a bilevel Stackelberg game, VMDNet exploits multi-periodic structures in demand series while avoiding information leakage. Experiments on three electricity demand datasets show that VMDNet consistently outperforms strong baselines across multiple forecasting horizons. Future work may explore multivariate forecasting through Multivariate Variational Mode Decomposition \cite{Rehman_2019}, as well as differentiable neural VMD variants for end-to-end gradient based optimisation.


\section*{Acknowledgment}
\noindent
The authors acknowledge the use of Isambard-AI National AI Research Resource (AIRR) funded by UKRI and DSIT [ST/AIRR/I-A-I/1023] and Isambard 3 Tier-2 HPC Facility funded by UKRI and EPSRC [EP/X039137/1].

\bibliographystyle{IEEEtran}
\bibliography{references}

\end{document}